# Improved Adaboost Algorithm for Web Advertisement Click Prediction Based on Long Short-Term Memory Networks

Qixuan Yu[1] , Xirui Tang[2]* , Feiyang Li[3] ,Zinan Cao[4] ,

[1]College of Computing, Georgia Institute of Technology, Atlanta, GA, 30332-0280, USA

[2]College of Computer Sciences, Northeastern University, Boston, MA, 02115, USA

[3]Department of Computer Science, University of Illinois Urbana-Champaign, Champaign, IL, 61820, USA

[4]Department of General Systems Studies, The University of Tokyo, Tokyo, 113-8654, Japan

* Corresponding author: e-mail: tang.xir@northeastern.edu

*Abstract*—This paper explores an improved Adaboost algorithm based on Long Short-Term Memory Networks (LSTMs), which aims to improve the prediction accuracy of user clicks on web page advertisements. By comparing it with several common machine learning algorithms, the paper analyses the advantages of the new model in ad click prediction. It is shown that the improved algorithm proposed in this paper performs well in user ad click prediction with an accuracy of 92%, which is an improvement of 13.6% compared to the highest of 78.4% among the other three base models. This significant improvement indicates that the algorithm is more capable of capturing user behavioural characteristics and time series patterns. In addition, this paper evaluates the model's performance on other performance metrics, including accuracy, recall, and F1 score. The results show that the improved Adaboost algorithm based on LSTM is significantly ahead of the traditional model in all these metrics, which further validates its effectiveness and superiority. Especially when facing complex and dynamically changing user behaviours, the model is able to better adapt and make accurate predictions. In order to ensure the practicality and reliability of the model, this study also focuses on the accuracy difference between the training set and the test set. After validation, the accuracy of the proposed model on these two datasets only differs by 1.7%, which is a small difference indicating that the model has good generalisation ability and can be effectively applied to real-world scenarios..

## I. INTRODUCTION

With the rapid development of the Internet, online advertising has become an important part of enterprise marketing. The click-through rate (CTR) of web advertisements is an important indicator to measure the effectiveness of advertisements, which is usually defined as the ratio of the number of times users click on an advertisement to the number of times the advertisement is displayed [1]. A high click-through rate means that an advertisement can effectively attract users' attention and increase brand exposure and conversion rate. Therefore, understanding and predicting the click-through rate is crucial for optimising advertisement placement strategies and improving ROI.

In recent years, with the surge of data volume and the improvement of computing power, researchers have begun to explore in depth the factors that affect the click-through rate of web advertisements [2]. These factors include user characteristics (e.g., age, gender, geographic location), ad content (e.g., copy, images, videos), display location, and time of day [3]. By analysing these variables, companies can better locate their target audience and thus develop more precise marketing strategies. However, due to the complexity and diversity of factors affecting the click-through rate, it is often difficult for traditional statistical methods to capture their intrinsic patterns, which provides a wide space for the application of machine learning techniques in this field [4].

Machine learning is a method of training models through data so that they can automatically identify patterns and make predictions. In web advertising click rate prediction, machine learning algorithms are able to process massive amounts of data, extract potential features from it, and make accurate predictions based on historical data. Common machine learning algorithms include linear regression [5], decision trees [6], random forests [7], support vector machines [8] and deep learning.

Linear regression models can be used to build simple and intuitive CTR prediction models by analysing the linear relationship between features and click behaviour. However, in practice, higher-level algorithms such as decision trees and random forests are widely used because of the complex non-linear relationships that may exist between features. These algorithms can better capture the effects of interactions between different features on click-through rates by constructing multi-layer decision rules.

Deep learning techniques have been rapidly developed in recent years, and they have achieved significant results in areas such as image recognition and natural language processing. In CTR prediction, deep neural networks (DNNs) are able to process high-dimensional features and extract deep information through multi-layer nonlinear transformations. This allows deep learning models to excel in handling sparse data with a large number of category variables (e.g., user IDs and product IDs). In addition, Convolutional Neural Networks (CNNs) and Recurrent Neural Networks (RNNs) are also used to analyse image and text information to further improve CTR prediction accuracy. In this paper, we select open-source datasets and improve the Adaboost algorithm using Long Short-Term Memory Networks for predicting users' clicks on web advertisements, and compare them using a variety of commonly used machine learning algorithms to analyse the advantages of this paper's algorithm in predicting users' advertisement clicks.

## II. INTRODUCTION TO THE DATASET

The dataset used in this paper is an open source dataset which can be used to predict whether a user clicks on an advertisement on a web page or not.The dataset contains several data features including time spent on site per day, age, region, income, daily internet usage, subject of advertisement, city, gender, country, time of day and whether or not they clicked on the advertisement.The whether or not they clicked on the advertisement is the user's behaviour, which is the target variable, 0 means did not click on the advert and 1 indicates that the user clicked on the advert. Some of the datasets were presented and the results are shown in Table 1.

TABLE I. SELECTED DATA SETS.

| Daily Time Spent on Site | Age | Area Income | Daily Internet Usage | Clicked on Ad |
|---|---|---|---|---|
| 62.26 | 32 | 69481.85 | 172.83 | 0 |
| 41.73 | 31 | 61840.26 | 207.17 | 0 |
| 44.4 | 30 | 57877.15 | 172.83 | 0 |
| 59.88 | 28 | 56180.93 | 207.17 | 0 |
| 49.21 | 30 | 54324.73 | 201.58 | 1 |
| 51.3 | 26 | 51463.17 | 131.68 | 0 |
| 66.08 | 43 | 73538.09 | 136.4 | 1 |
| 36.08 | 26 | 74903.41 | 228.78 | 0 |
| 46.14 | 33 | 43974.49 | 196.77 | 0 |
| 51.65 | 51 | 74535.94 | 188.56 | 0 |
| 47.64 | 29 | 53431.35 | 200.71 | 0 |
| 65.07 | 34 | 34191.23 | 187.09 | 1 |
| 55.6 | 24 | 52252.91 | 167.22 | 1 |
| 62.26 | 25 | 50671.6 | 138.71 | 0 |
| 78.84 | 27 | 69646.35 | 239.32 | 0 |
| 56.39 | 40 | 40468.53 | 140.46 | 1 |
| 54.43 | 39 | 56180.93 | 124.44 | 0 |

We statistically analysed the quantitative data in the dataset, counting the maximum, minimum, median, variance and mean of each variable, and the results are shown in Table 2.

TABLE II. STATISTICAL ANALYSIS OF DATA.

| Variable Name | Maximum | Minimum | Mean | Median |
|---|---|---|---|---|
| Daily Time Spent on Site | 90.97 | 32.6 | 61.661 | 59.59 |
| Age | 60 | 19 | 35.94 | 35 |
| Area Income | 79332.33 | 13996.5 | 53840.048 | 56180.93 |
| Daily Internet Usage | 269.96 | 105.22 | 177.76 | 178.92 |
| Clicked on Ad | 1 | 0 | 0.492 | 0 |

## III. METHOD

Before proceeding to the main experiment, we first predict user clicks using three commonly used machine learning algorithms, i.e., decision trees, random forests, and the XGBoost algorithm.

### A. Decision tree

Decision tree is a supervised learning algorithm based on tree structure for classification and regression tasks. It forms a tree by recursively dividing the dataset into different subsets. Each internal node represents a feature (attribute), each branch represents some value of that feature, and each leaf node represents the final output. The structure of a decision tree is shown schematically in Fig. 1.

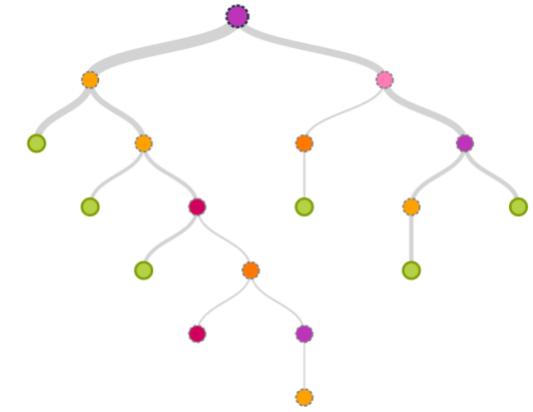

Figure 1. The structure of a decision tree.

### B. Random forest

Random forest is a method of integrated learning that improves the accuracy and robustness of a model by constructing multiple decision trees and combining their predictions. In the training process, Random Forest randomly draws multiple samples from the original dataset and trains multiple decision trees on these samples. At the same time, only some of the randomly selected features are considered in the feature selection at each node. This "random" strategy helps to reduce overfitting and improve the generalisation ability of the model. Ultimately, the random forest generates the final prediction by voting (classification) or averaging (regression) [9].

### C. XGBoost

XGBoost is an efficient and flexible gradient boosting framework that is widely used in machine learning competitions and practical applications. Unlike traditional boosting methods, XGBoost introduces a regularisation term to control the model complexity and thus reduce overfitting. In addition, it employs second-order gradient information, which makes the optimisation process more efficient.XGBoost supports parallel computing, can handle large-scale datasets, and provides a variety of features, such as missing-value processing, cross-validation, etc., which enable it to perform well in handling complex tasks.

### D. Long Short-Term Memory Networks

Long Short-Term Memory Network (LSTM) is a special type of Recurrent Neural Network (RNN) that is specifically designed to process and predict time series data. Traditional RNNs are prone to the problem of vanishing or exploding gradients when dealing with long sequences, resulting in models that are unable to effectively learn long-term dependencies. LSTM solves this problem by introducing "memory units" and three gating mechanisms (input gate, forget gate and output gate), which can better capture the long-term dependency features in time series. The structure of the LSTM algorithm is shown in Figure 2.

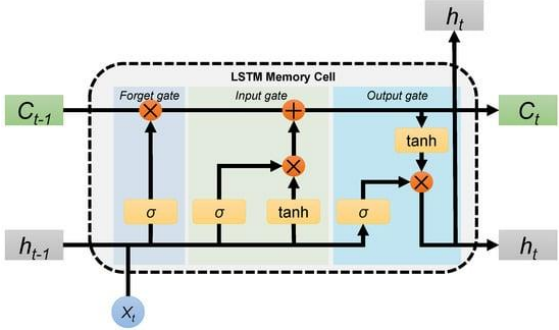

Figure 2. The structure of the LSTM algorithm.

At the heart of the LSTM is its unique structure, which contains a cell state and three important gates: the input gate determines what information can be stored into the cell state, the forgetting gate controls what information needs to be discarded from the cell state, and the output gate determines the effect of the current cell state on the next layer. These gates generate a value between 0 and 1 indicating the importance of the information through a sigmoid activation function. Specifically, input gates update the cell state based on the current input and the hidden state of the previous moment; forgetting gates determine whether the previous information needs to be retained based on the same information; and output gates convert the cell state into a final output that can be passed on to the next time step or the next network layer [10].

### E. Adaboost.

Adaboost is an integrated learning algorithm designed to improve the predictive performance of a model by combining multiple weak classifiers. The basic idea is to combine multiple underperforming classifiers (weak classifiers) into a single strong classifier. Adaboost trains each weak classifier iteratively and dynamically adjusts the sample weights based on its performance on the training set, thus making subsequent classifiers pay more attention to those samples that were misclassified by the previous classifier. This weighting mechanism makes the final model better able to capture complex patterns in the data. The model principle of Adaboost is shown in Fig. 3.

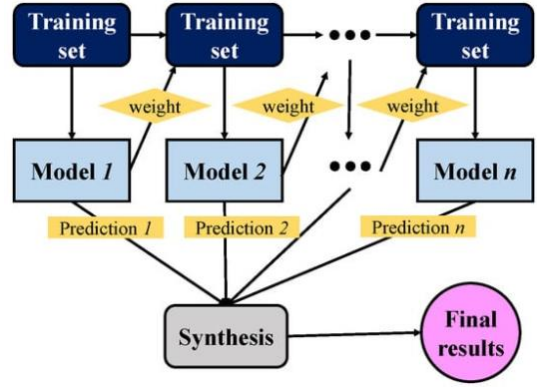

Figure 3. The model principle of Adaboost .

In detail, Adaboost first initialises the weights of each sample so that all samples have equal weights. Then, it will iteratively perform the following steps: first train a weak classifier and calculate the error rate of this classifier on the training set. Next, the sample weights are updated based on the error rate of that classifier such that the weights of the misclassified samples are increased while the weights of the correctly classified samples are decreased. In this way, in the next round of training, the new weak classifiers will pay more attention to the samples that are difficult to classify. Each weak classifier is assigned a weight value that is proportional to its accuracy, and the final model is the result of weighting all weak classifiers.

### F. Optimising Adaboost's classification algorithm based on long short-term memory networks

Combining LSTM with Adaboost allows the powerful sequence modelling capability of LSTM to be used to generate base learners, while Adaboost is used to enhance the performance of these learners. Specifically, the time series data is first modelled using LSTM to extract effective features, which are then fed into Adaboost to train multiple weak classifiers. Each weak classifier can be an LSTM model based on different parameters or structures, so that the diversity and complexity in the data can be fully exploited. In this way, Adaboost not only improves the robustness of the model, but also enhances the resistance to noise and outliers, thus improving the prediction accuracy.

In the implementation process, the data first needs to be preprocessed, including steps such as normalisation and splitting into training and test sets. Next, the LSTM model is constructed and trained to obtain preliminary prediction results. Then, the error of each sample is calculated based on the prediction results and the sample weights are adjusted to provide a basis for subsequent training. This process is repeated until the set number of iterations is reached or the error meets the requirements. In the final stage, all weak classifiers are combined to form a strong classifier for prediction on new data.

Combining the long and short-term memory network with the Adaboost algorithm is an innovative and effective method, which organically integrates the advantages of sequence modelling in deep learning with the powerful performance enhancement ability in integrated learning, providing a new way of thinking for solving complex problems.

## IV. EXPERIMENTMENT

In terms of dataset division, this paper divides the dataset according to the ratio of 7:3, uses 70% of the data to train the model, uses 30% of the data to test the trained model, outputs the confusion matrix of the test set, and at the same time, outputs model evaluation parameters such as precision, collinearity, recall, and F1 score of the test set.

The confusion matrices of the three sets of basic experiments are output as shown in Fig. 4, Fig. 5 and Fig. 6. The model evaluation parameters for the training and test sets of the three sets of base experiments are also output, as shown in Table 3, Table 4 and Table 5.

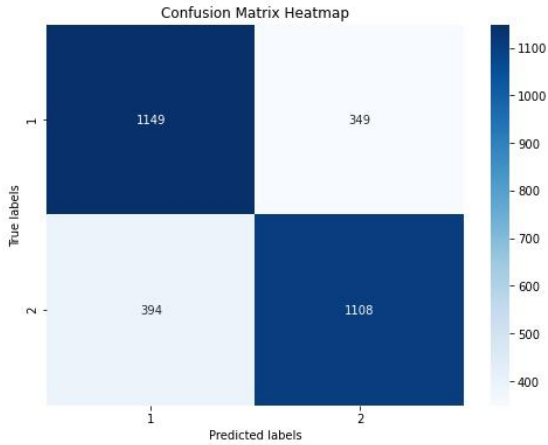

Figure 4.   Confusion matrix.

TABLE III.     EVALUATION METRICS FOR TRAINING AND TEST SETS.

|  | Accuracy | Recall | Precision | F1 |
|---|---|---|---|---|
| Training Set | 0.778 | 0.778 | 0.778 | 0.778 |
| Test set | 0.752 | 0.752 | 0.753 | 0.752 |

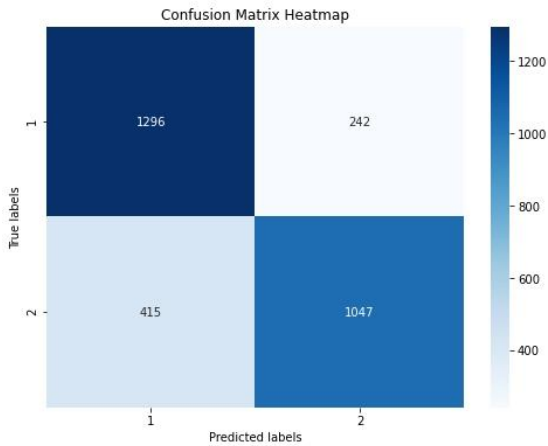

Figure 5.   Confusion matrix.

TABLE IV.     EVALUATION METRICS FOR TRAINING AND TEST SETS.

|  | Accuracy | Recall | Precision | F1 |
|---|---|---|---|---|
| Training Set | 0.786 | 0.786 | 0.789 | 0.785 |
| Test set | 0.781 | 0.781 | 0.784 | 0.78 |

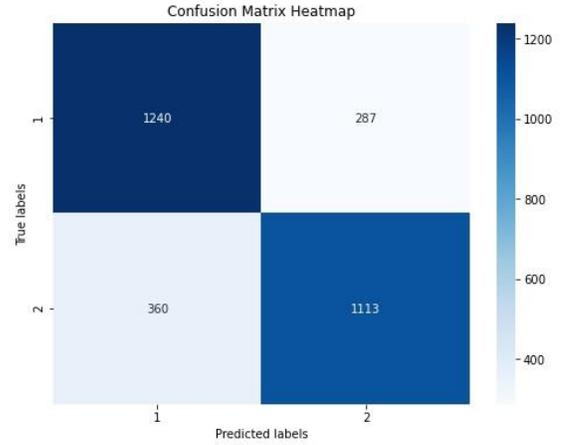

Figure 6.   Confusion matrix.

TABLE V.     EVALUATION METRICS FOR TRAINING AND TEST SETS.

|  | Accuracy | Recall | Precision | F1 |
|---|---|---|---|---|
| Training Set | 0.937 | 0.937 | 0.937 | 0.937 |
| Test set | 0.784 | 0.784 | 0.785 | 0.784 |

The confusion matrix of this paper's method and the model evaluation parameters for the training and test sets are output, as shown in Figure 7 and Table 6.

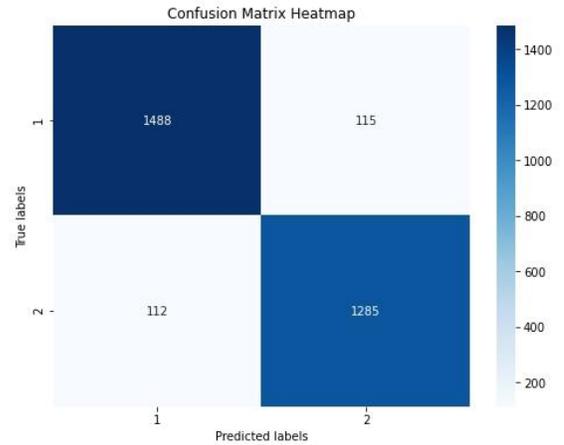

Figure 7.   Confusion matrix.

TABLE VI.     EVALUATION METRICS FOR TRAINING AND TEST SETS.

|  | Accuracy | Recall | Precision | F1 |
|---|---|---|---|---|
| Training Set | 0.937 | 0.937 | 0.937 | 0.937 |
| Test set | 0.92 | 0.92 | 0.92 | 0.92 |

The individual models were compared and the histograms of the evaluation indicators were output, and the results are shown in Figure 8.

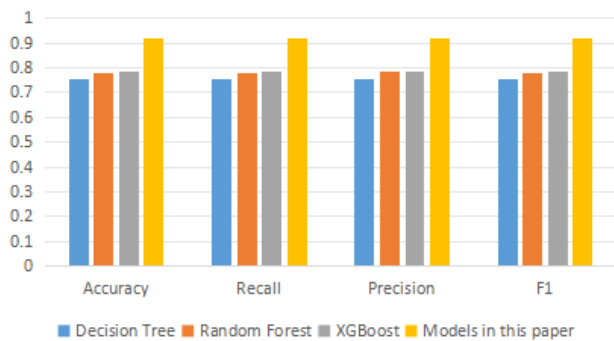

Figure 8.  Comparison of model evaluation indicators.

From the comparison results, it can be seen that the highest accuracy in user advertisement click prediction is the improved Adaboost algorithm based on long and short-term memory network proposed in this paper, with an accuracy of 92%, which is 13.6% higher than the highest accuracy of 78.4% among the three models, and the prediction accuracy is greatly improved. Meanwhile, the improved Adaboost algorithm based on long and short-term memory networks proposed in this paper is also significantly ahead of the other three base models in terms of accuracy, recall and F1 score. The accuracy of the model proposed in this paper differs by 1.7% between the training set and the test set, which is not a big difference, indicating that the generalisation ability of the model is better.

## V. Conclusion

In this paper, we adopt the Long Short-Term Memory (LSTM) network to make innovative improvements to the Adaboost algorithm, aiming to enhance the prediction of user clicks on web advertisements. By comparing with several common machine learning algorithms, we systematically analyse the advantages of the proposed algorithm in advertisement click prediction. The experimental results show that the Adaboost algorithm, which is improved based on LSTM, performs outstandingly in terms of accuracy, reaching 92%. This result is not only significantly higher than the highest accuracy rate of 78.4% in the traditional model, but also improves by 13.6%, showing the powerful ability of the model in processing user behaviour data.

Further analysis reveals that the model proposed in this paper also outperforms the other three base models in several performance metrics, including accuracy, recall, and F1 score. The improvement in these metrics implies that our model is not only able to identify positive examples (i.e., users clicking on advertisements) more accurately, but also effectively reduces the incidence of false-positives and false-negatives, which improves the overall prediction quality. In addition, the difference in accuracy between the training and test sets is only 1.7%, which indicates that the proposed model has good generalisation ability and can effectively adapt to new data without overfitting the training set.

In summary, by combining LSTM with Adaboost, we have implemented a new advertisement click prediction method that excels in both accuracy and stability. This research result not only provides strong data support for advertisement placement strategies, but also lays the foundation for future research in related fields. We believe that as the amount of data continues to increase and computing power improves, deep learning-based methods will show greater potential in more practical applications, bringing more accurate and efficient solutions to the industry.